\documentclass[11pt]{article}

\usepackage{macros/packages}
\usepackage{macros/editing-macros}
\usepackage{macros/formatting}
\usepackage{macros/statistics-macros}

\definecolor{lightblue}{rgb}{0.2, 0.4, 0.8}
\definecolor{yellowPPT}{RGB}{255,245,217}
\definecolor{greenPPT}{RGB}{197,224,180}
\definecolor{orangePPT}{RGB}{240,148,116}
\definecolor{bluePPT}{RGB}{180,199,231}

\newcommand{\githubicon}{\faGithub}
\newcommand{\codeicon}[1]{\href{#1}{\githubicon}}

\newcommand{\No}{%
\tikz[scale=0.23] {
    \draw[red!70!black,thick,line width=0.7,line cap=round] (0,0) to [bend left=6] (1,1);
    \draw[red!70!black,thick,line width=0.7,line cap=round] (0.2,0.95) to [bend right=3] (0.8,0.05);
}}
\newcommand{\Yes}{%
\tikz[scale=0.23] {
    \draw[green!60!black,thick,line width=0.7,line cap=round] (0.25,0) to [bend left=10] (1,1);
    \draw[green!60!black,thick,line width=0.8,line cap=round] (0,0.35) to [bend right=1] (0.23,0);
}}
\newcommand{\HalfYesNo}{%
  \tikz[scale=0.23]{
    \draw[yellow!60!black,thick,line width=0.7,line cap=round] (0.25,0) to [bend left=10] (1,1);
    \draw[yellow!60!black,thick,line width=0.8,line cap=round] (0,0.35) to [bend right=1] (0.23,0);
    \draw[yellow!70!black,thick,line width=0.7,line cap=round] (0.2,0.95) to [bend right=3] (0.8,0.05);
  }%
}

\lstdefinelanguage{json}{
  basicstyle=\ttfamily\footnotesize,
  numbers=left,
  numberstyle=\tiny\color{gray},
  stepnumber=1,
  numbersep=8pt,
  showstringspaces=false,
  breaklines=true,
  frame=single,
  backgroundcolor=\color{black!5},
  literate=
   *{0}{{{\color{blue}0}}}{1}
    {1}{{{\color{blue}1}}}{1}
    {2}{{{\color{blue}2}}}{1}
    {3}{{{\color{blue}3}}}{1}
    {4}{{{\color{blue}4}}}{1}
    {5}{{{\color{blue}5}}}{1}
    {6}{{{\color{blue}6}}}{1}
    {7}{{{\color{blue}7}}}{1}
    {8}{{{\color{blue}8}}}{1}
    {9}{{{\color{blue}9}}}{1}
    {:}{{{\color{red}:}}}{1}
    {,}{{{\color{red},}}}{1}
    {"}{{{\color{orange}"}}}{1},
}

\def\addauthnote#1#2{
    \expandafter\def\csname#1\endcsname##1{\todo[inline,size=\footnotesize,color=#2]{\textbf{\underline{\texttt{#1}}:} ##1}\xspace}
}
\addauthnote{naimeng}{blue!25}
\addauthnote{tommaso}{green!25}

\newcommand{\showcomments}{1}   

\newcommand{\tc}[1]{%
  \ifnum\showcomments=1%
    \textcolor{green}{#1}%
  \fi%
}

\newcommand{\minisection}[1]{%
  \ifnum\showcomments=1%
    \textcolor{purple}{\textit{#1.}}\quad%
  \fi%
}

\newcommand{\titleFont}{\bfseries\fontfamily{qpl}\selectfont}
\newcommand{\authorFont}{\fontfamily{qpl}\selectfont}

\definecolor{boxfill}{RGB}{244,246,254}
\definecolor{boxaccent}{RGB}{40, 55, 100}
\definecolor{titleblue}{HTML}{1a2a6c}

\newenvironment{ack}{\paragraph{Acknowledgments.}}{}

\begin{document}

\abovedisplayskip=8pt plus0pt minus3pt
\belowdisplayskip=8pt plus0pt minus3pt

\begin{tcolorbox}[
  enhanced,
  colback=boxfill,
  colframe=boxaccent,
  boxrule=0.6pt,
  arc=6pt,
  left=18pt, right=18pt, top=14pt, bottom=12pt,
  breakable,
]
\begingroup
\centering
{\LARGE\titleFont\color{titleblue}
SynthTools: A Framework for Scaling \\ Synthetic Tools for Agent Development  \par}
\vspace{0.8em}
{\normalsize \authorFont
\textbf{Tommaso Castellani}\textsuperscript{*\,1}\quad
\textbf{Naimeng Ye}\textsuperscript{*\,1}\quad
\textbf{Daksh Mittal}\textsuperscript{*\,1}\quad
\textbf{Thomson Yen}\textsuperscript{*\,1}\\[3pt]
\textbf{Emmanouil Koukoumidis}\textsuperscript{2}\quad
\textbf{William Zeng}\textsuperscript{2}\quad
\textbf{Hongseok Namkoong}\textsuperscript{1}\par}
\vspace{0.8em}
{\normalsize \authorFont
\textsuperscript{1}Columbia University\quad
\textsuperscript{2}Oumi\par}
\vspace{0.5em}
{\normalsize\authorFont\hypersetup{urlcolor=titleblue}\href{https://github.com/namkoong-lab/SynthTools}{\faGithub} \hspace{0.3em} $|$ \hspace{0.3em} \href{https://huggingface.co/collections/namkoong-lab/synthtools}{\raisebox{-0.2em}{\includegraphics[height=1em]{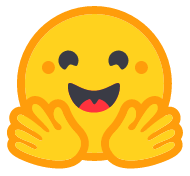}}}}\\[0.5em]
\endgroup
{\color{boxaccent!50}\hrule height 0.3pt}
\vspace{0.7em}
\begingroup
\authorFont
For agentic systems to use external tools to solve complex, long-horizon tasks, we need a large set of diverse and controllable tool-use environments. We introduce \textit{SynthTools}, a fully LLM-based pipeline spanning the entire lifecycle: environment generation, simulation, validation and task construction. By operating end-to-end through LLMs, our framework complements other tool-use environments bottlenecked by the complexity of real APIs, and ensures scalability and controllability by design. The framework consists of three components: \textit{top-down environment generation}, which hierarchically constructs diverse, domain-grounded tool environments; \textit{environment simulation and validation}, which ensures tools can be reliably emulated and filters out those that cannot; and \textit{bottom-up task and trajectory generation}, which produces solvable and verifiable tasks together with multi-step trajectories, exposing control over difficulty, length, trajectory composition, and domain focus to guarantee flexibility. As a concrete instantiation, we release the dataset comprising $73{,}883$ validated tools across $6{,}800$ environments and $100$ fields, $79{,}925$ verifiable tasks as well as the pipeline to generate trajectories at scale. Training Qwen3 models of various sizes on a corpus of trajectories generated from these tasks yields gains across multiple tool-use benchmarks, including real APIs, indicating tool-use capabilities trained on synthetic data may transfer to some real environments. Together, these results suggest that SynthTools can serve as a useful infrastructure for large-scale training of tool-use agents.
\par
\endgroup
\vspace{0.7em}
{\centering\small\authorFont \textbf{Correspondence:} \texttt{tc3527@columbia.edu}
\qquad \qquad\qquad \qquad \qquad \qquad \qquad \textsuperscript{*}Equal contribution}
\vspace{0.3em}
\end{tcolorbox}

\vspace{1.0em}


\section{Introduction}
\label{sec:introduction}

Large language model (LLM) agents have drawn considerable attention for their potential to tackle complex, real-world tasks. 
To complete such tasks, modern agents are increasingly envisioned to make coordinated and precise use of multiple tools \citep{xu2023tool, qin2023toolllm, barres2025tau2benchevaluatingconversationalagents}. 

\begin{table}
\centering
\fontsize{8.7}{9}\selectfont
\setlength{\tabcolsep}{1.8pt}
\renewcommand{\arraystretch}{1.2}
\begin{tabular}{@{}l*{5}{c}@{}}
\toprule
& \makecell{\textbf{Simulation}} 
& \makecell{\textbf{Generated /}\\\textbf{Curated}} 
& \makecell{\textbf{Field}\\\textbf{Scalability}} 
& \makecell{\textbf{Tool}\\\textbf{Scalability}} 
& \makecell{\textbf{Complex}\\\textbf{Tools}} \\
\midrule
\textbf{SynthTools}   & LLM          & Generated  & \Yes        & \Yes        & \Yes \\
$\tau^2$-Bench           & Programmatic & Curated    & \No         & \No         & \Yes  \\
ACEBench               & Programmatic & Both    & \No         & \No         & \No  \\
ToolLLM              & Programmatic & Real APIs  & \No         & \No       & \Yes  \\
StableToolBench        & Both         & Real APIs  & \No         & \No         & \Yes \\
Simia                  & LLM          & Both  & \No         & \No         & \Yes \\
ToolAlpaca             & LLM          & Both  & \HalfYesNo  & \HalfYesNo  & \No  \\
RandomWorld            & Programmatic & Generated  & \No         & \Yes        & \No  \\
AWM                    & Programmatic & Generated  & \Yes        & \Yes        & \No  \\
EnvScaler              & Programmatic & Generated  & \HalfYesNo  & \No         & \No  \\
\bottomrule
\end{tabular}
\caption{Comparison of tool-use pipelines. LLM-based simulation enables richer, more complex tools but lacks deterministic execution; programmatic simulation is deterministic but restricts tool complexity. Curated suites and real APIs offer authenticity but limit scale, while generated environments scale further.}
\label{tab:comparison}
\end{table}
However, current models continue to face challenges in using tools reliably \citep{barres2025tau2benchevaluatingconversationalagents, wang2024gta, patil2025bfcl}. 
As evident throughout major advances in machine learning \citep{Deng_Imagenet,Hoffmann_Chinchilla}, the scale and diversity of training sets are among the most critical factors influencing the quality of a model.
Therefore, to realize the vision of autonomous tool-use agents, it is essential to obtain a comprehensive set of tool-use training environments.

Tool-use pipelines vary along two largely independent axes: how environments are \textit{generated} and how tool calls are \textit{simulated} at runtime. A straightforward approach to environment generation is to directly use real-world APIs as tools and construct tasks from them, since they offer authenticity and reflect practical constraints \citep{qin2023toolllm, fang2025towards, prabhakar2025apigen, patil2024gorilla, lu2026youtullmunlockingnativeagentic}. However, this choice bounds how these environments can be simulated: real APIs pose practical problems during interaction such as requirements for API keys, usage quotas, rate limits, and frequent interface changes or deprecations, all of which complicate scalable environment creation \citep{guo_stabletoolbench}. One option to compromise on this is to retain real API documentation but simulates tool behavior using an LLM, decoupling generation from simulation  \cite{li2025simulatingenvironmentsreasoningmodels}. An alternative solution is to manually curate tool suites or build high-fidelity replicas \citep{xu2023tool, barres2025tau2benchevaluatingconversationalagents,  patil2025bfcl, li_apibank, chen2025acebenchwinsmatchpoint}, paired with programmatic simulation in separated sandboxes. These hand-crafted collections are naturally limited in both the number of available tools and the diversity of covered domains; for instance, ACEbench \citep{chen2025acebenchwinsmatchpoint} includes only eight broad domains, and $\tau^2$-bench \citep{barres2025tau2benchevaluatingconversationalagents} covers just five, making them insufficient as exhaustive training environments. 

For the purpose of scaling beyond real and curated APIs, synthetic generation has emerged as a major paradigm, offering a flexible and controllable solution that addresses both the unreliability of real APIs and the limited scale of hand-crafted surrogates. Within this paradigm, a key distinction emerges in how simulation is handled: the vast majority of these works generate environments whose tools are then executed programmatically \citep{song2026envscalerscalingtoolinteractiveenvironments, wang2026agentworldmodelinfinity, sullivan2025proceduralenvironmentgenerationtooluse, prabhakar2025apigen, ye2026feedbackdriventooluseimprovementslarge, fang2025towards, patil2024gorilla, zhuang2023toolqa, wang2023mint, guo_stabletoolbench, hafner2025trainingagentsinsidescalable}, while LLM-based simulation remains largely underexplored, limited to  using or evolving real API documentation \cite{li2025simulatingenvironmentsreasoningmodels, guo_stabletoolbench, tang2023toolalpaca}.

These collective efforts have produced a landscape of diverse pipelines, each shaped by design choices that make it well suited for particular use cases. Curated and API-grounded environments, for instance, are naturally suited for evaluation, where reproducibility and ground-truth verification are paramount \citep{xu2023tool, barres2025tau2benchevaluatingconversationalagents, chen2025acebenchwinsmatchpoint}. For training, however, the binding constraints are scale and domain diversity, which existing collections struggle to provide: prominent tool-use corpora remain either closed-source \citep{kimik2} or limited in the number of domains and tools they cover. This gap is increasingly salient given recent evidence that an intermediate \emph{midtraining} phase, positioned between pretraining and task-specific fine-tuning, can endow models with higher-level capabilities such as reasoning, tool use, and instruction following~\citep{ tongyideepresearchteam2025tongyideepresearchtechnicalreport, lu2026youtullmunlockingnativeagentic, liu2026midtrainingbridgespretrainingposttraining}, creating the need for pipelines capable of generating large-scale, diverse datasets.

Furthermore, even with a large and well-designed tool set, the problem is far from solved. Constructing realistic, non-trivial tasks remains a fundamental bottleneck \citep{sheng2024verl, andrews2025arescalingagentenvironments, griggs2025skrylv01, spisak2025openenv}. Meaningful tasks require structured workflows in which multiple tools must be used in coordination; for example, processing a product return by interacting with tools such as \texttt{ReturnStatusTracker} and \texttt{CustomerNotifier} (see Section~\ref{sec:dataset}). Naively prompting LLMs to generate such tasks often yields shallow or implausible scenarios that do not demand long-horizon reasoning, coordinated tool use, or decision-making under uncertainty. Moreover, task success is difficult to evaluate systematically: many benchmarks rely solely on LLM-based judges \citep{guo_stabletoolbench}, while others require manual verification \citep{chen2025acebenchwinsmatchpoint}, limiting scalability. As a result, the field still lacks a scalable and principled way to generate both rich tool ecosystems and the verifiable tasks needed for agent training and evaluation.

In this work, we propose \textbf{SynthTools}, a fully automated yet structured pipeline designed to complete the current agentic training landscape. By being LLM-simulated end-to-end, it reaches scales and domain coverage that real-API, curated, and programmatic pipelines cannot, filling the gap open for large-scale, diverse training. The framework prioritizes scalability and controllability, supporting systematic control over task difficulty and domain coverage to serve as a practical training resource for tool-use agents. As a concrete instantiation, we release a dataset of $73{,}883$ validated tools across $6{,}800$ environments and $79{,}925$ verifiable tasks; over $2\times$ larger than prior collections in both number of domains and tools per domain.

\begin{figure*}
    \centering
    \includegraphics[width=\textwidth]{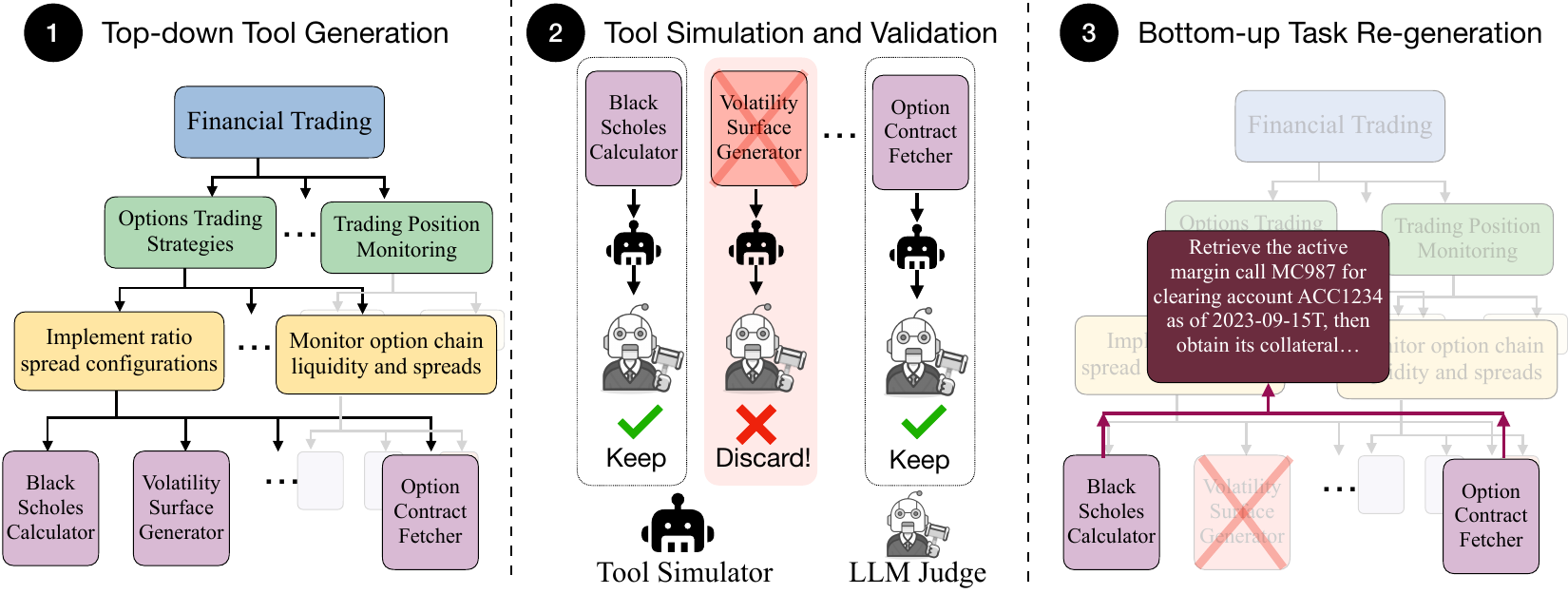}
    \caption{Illustration of the \textit{SynthTools} pipeline.}
    \label{fig:tool_gen}
\end{figure*}

\section{SynthTools Framework and Capabilities}
SynthTools integrates three core modules designed with specific desiderata to ensure scalability, realism, and reliability across the entire pipeline: \begin{enumerate}[leftmargin=*,itemsep=2pt,topsep=2pt,parsep=0pt]
    \item \textbf{Top-down Environment Generation.} This module hierarchically constructs domain-grounded tools as opposed to a flat list of APIs. It \emph{scalably} generates \emph{diverse} tools that mirror the complexity of real-world APIs (Figure~\ref{fig:example-tool}). We concretely instantiate a dataset of $73{,}883$ tools that exceeds prior work by more than \textbf{2$\times$} in both the number of domains and tools per domain, with capacity for at least an order of magnitude further. Moreover, SynthTools covers richer domains (e.g., \texttt{Financial Trading}, \texttt{Transportation Logistics}) and more complex tools than existing curated~\citep{tang2023toolalpaca, barres2025tau2benchevaluatingconversationalagents, chen2025acebenchwinsmatchpoint} and generated environments~\citep{song2026envscalerscalingtoolinteractiveenvironments, wang2026agentworldmodelinfinity, sullivan2025proceduralenvironmentgenerationtooluse, prabhakar2025apigen, ye2026feedbackdriventooluseimprovementslarge, fang2025towards, patil2024gorilla, tang2023toolalpaca, zhuang2023toolqa, wang2023mint, guo_stabletoolbench, hafner2025trainingagentsinsidescalable}, including more diverse tools with respect to programmatic environments that restrict the scope to create, read, update, and delete operations~\citep{wang2026agentworldmodelinfinity}. The hierarchical design also provides tunability: practitioners can select specific fields or subdomains to tailor the generated toolset to their application of interest. Figure~\ref{fig:example-tool} and Table~\ref{tab:comparison} illustrate the tool complexity and qualitative advantages.

    \item \textbf{Environment Simulation and Validation}. This module ensures generated tools can be reliably emulated. It consists of two submodules. \textit{Simulation} takes in tool specifications and produces consistent and flexible responses (e.g., returning \texttt{`no seat available'} for a fully booked flight). We achieve \textbf{94\%} accuracy in simulating tool responses. 
    \textit{Validation} verifies response correctness, filters unstable tools, and supports retries to improve reliability. Prioritizing a low false positive rate, it achieves \textbf{97\%} accuracy in identifying incorrect behaviors. 
    \item \textbf{Bottom-up Task and Trajectory Generation}. Complementing the top-down construction of tools, this module builds tasks back up from the validated toolset. Each tool defines an atomic capability that gives rise to a mini-task along with its corresponding ground truth tool call. We take the tools generated from the same task in module 1, and assemble them into a coherent, multi-step workflow. Because each step is paired with a known correct tool call, composing them yields tasks with explicit ground-truth solution trajectories. This pipeline is partially inspired by AgentSynth \cite{xie2025agentsynthscalabletaskgeneration} in tasks for computer-use, and an example is given by Figure~\ref{fig:example-task-1}. Using this pipeline, we construct $79{,}925$ tasks across $100$ fields.

\end{enumerate}

Using this bidirectional pipeline, the resulting task collection spans diverse domains and difficulty levels, requiring coordinated multi-step tool use and long-horizon reasoning (see Figure~\ref{fig:example-task-1}). We present a case study showing that midtraining on this dataset improves Qwen3's agentic tool-use capabilities, with consistent gains across multiple benchmarks.

\section{Top-down Environment Generation}
\label{sec:tool-generation}
We use LLMs to synthesize tool environments, but naive prompting produces redundant and overly simplistic APIs. For example, directly prompting ChatGPT-5~Pro~\cite{singh2026openaigpt5card} often yields trivial tools such as \texttt{robotics.create\_task} or \texttt{docs.delete}, which lack the granularity needed for meaningful agent tasks (Figure~\ref{fig:example-tool}). Such shallow tools are ill-suited for studying complex tool-use behavior.

To generate tools with sufficient \textit{diversity} and \textit{complexity}, we propose a hierarchical domain evolution procedure (Figure~\ref{fig:tool_gen}) that refines a high-level domain into a coherent toolset. Starting from a domain, we progressively decompose it into subdomains, task families, and concrete tools whose interfaces encode domain-specific constraints and interactions. LLM generation allows for tailored complexity of the tool schema by choosing the number of arguments and the type where lists and complex object types are harder than string and numbers. 

Concretely, the module proceeds as follows:
\begin{enumerate}[leftmargin=*,itemsep=2pt,topsep=2pt,parsep=0pt]
\item \textbf{Field $\rightarrow$ Sub-domain.}
    Given a seed set of fields (e.g., \textit{healthcare}, \textit{finance}, \textit{materials science}), we prompt a large language model (LLM) to propose coherent subdomains that: (a) partition typical workflows; (b) surface stakeholders and entities operating in the field; and (c) admit meaningful, tool-addressable operations.
    \item \textbf{Sub-domain $\rightarrow$ Task.}
For each subdomain, the model proposes a family of tasks, each with a natural task description suitable for documentation and test-case generation.
    \item \textbf{Task $\rightarrow$ Tool.}
    The model then instantiates each task with concrete tools.
    Tools are encouraged to be \emph{composable}: each specifies upstream dependencies (what it consumes) and downstream affordances (what it produces), enabling tasks that require nested tool use.
\end{enumerate}
Targeted prompting at each stage controls diversity, complexity, and input/output (I/O) behavior. The module outputs \textit{tools}, each defined by (i) a name and natural-language description, (ii) a parameter schema, and (iii) an I/O contract (preconditions, postconditions, and error modes). Formally, a tool is a tuple (\texttt{name},\ \texttt{description},\ \texttt{parameters},\ \texttt{usage},\ \texttt{failure\_modes},\ \texttt{output\_schema}).

\begin{figure}
{\small
\begin{tcolorbox}[colback=orangePPT!5!white,
                  colframe=orangePPT,
                  title=Field to Tool Evolution,
                  fonttitle=\bfseries,
                  top=2pt, bottom=2pt, left=4pt, right=4pt,
                  boxsep=2pt]
\textbf{Field:} Financial Trading $\rightarrow$ \textbf{Sub Domain}: Options Trading Strategies $\rightarrow$ \textbf{Task:} Calculate implied volatility for specific option contracts $\rightarrow$ \textbf{Tool:} Black Scholes Calculator
\end{tcolorbox}
\begin{tcolorbox}[colback=orangePPT!5!white,
                  colframe=orangePPT,
                  title= The Generated Tool: Black Scholes Calculator,
                  fonttitle=\bfseries,
                  top=2pt, bottom=2pt, left=4pt, right=4pt,
                  boxsep=2pt]
\textbf{Description:}
Calculates option prices using the Black–Scholes–Merton model. Can price European call and put options based on inputs like underlying price, strike, time to expiration, risk-free rate, and volatility.

\textbf{Parameters:}
\begin{itemize}[leftmargin=*, topsep=0pt, itemsep=0pt, parsep=0pt]
  \item \textbf{option\_type} (string, required) – Type: \texttt{call} or \texttt{put}
  \item \textbf{underlying\_price} (number, required) – Current price of the underlying asset (\(S\))
  \item \textbf{...}
\end{itemize}

\textbf{Error Messages:}
\begin{itemize}[leftmargin=*, topsep=0pt, itemsep=0pt, parsep=0pt]
  \item Invalid option type: \texttt{option\_type} must be either \texttt{call} or \texttt{put}.
  \item \textbf{...}
\end{itemize}

\textbf{Usage:}
Provide \texttt{option\_type}, ..., to calculate the theoretical option price. Optionally include \texttt{dividend\_yield} for dividend-paying assets.

\textbf{Output Details:}
\begin{itemize}[leftmargin=*, topsep=0pt, itemsep=0pt, parsep=0pt]
  \item \textbf{option\_price} (number) – Calculated theoretical price of the option
  \item \textbf{...}
\end{itemize}
\end{tcolorbox}
}
\caption{An example of a tool generated through our pipeline for the \texttt{Financial Trading} field. The complete tool description is in Appendix~\ref{sec:example_tools}.}
\label{fig:example-tool}
\end{figure}
Owing to the generative flexibility of LLMs, our framework places no inherent limit on the number of fields or tools that can be produced. This allows the ecosystem to scale to meet the needs for given specific applications. As a concrete instantiation, we generate $100$ distinct fields, $6,800$ environments, with roughly $10-14$ tools. This results in a total of $85,447$ tools before filtering, and $73,883$ tools after filtering. This represents more than a $2\times$ increase in scale over existing datasets, with the potential to expand further if needed.

Beyond scale, another key characteristic is \textit{diversity}, which we demonstrate along two dimensions:
\begin{itemize}[leftmargin=*,itemsep=2pt,topsep=2pt,parsep=0pt]
    \item \textbf{Across fields:} we take 200 tools each from 20 different fields and observe sufficient breadth and diversity, see Figure~\ref{fig:tools_embedding_dist}. The full 100 field figure is in Appendix \ref{sec:dataset_details}. While the selected domains are not exhaustive, this experiment shows that our pipeline can produce a numerous and non-overlapping set of tools from various application areas.
    \item \textbf{Within field:} to test that scaling tool count yields genuinely novel tools, we generate up to 1{,}000 tools in an example e-commerce and retail domain. As shown in Figure~\ref{fig:tools_embedding_dist}, the distribution indicates high tool uniqueness. Generating 200 tools across several other domains yields consistent results, confirming that our pipeline can scale tools within a single field without significant redundancy.
\end{itemize}
Notably, because tasks are constructed exclusively from within-environment tools to guarantee a nested, interdependent structure, each environment is self-contained and deduplication is not a required preprocessing step. Additionally, we compute pairwise embedding similarities across all generated tools and find that only 9\% fall above a similarity threshold of $\tau = 0.85$, confirming that the hierarchical generation process naturally produces diverse, non-redundant tools even at scale. Further details are provided in Appendix~\ref{sec:deduplication_details}.

Moreover, we note that the generated tools are \textit{realistic}.
Figure~\ref{fig:example-tool} illustrates a tool generated by our pipeline in the Financial Trading domain.
The tools consist of rich failure modes, and output schemas, reflecting the diversity and complexity in real-world tool ecosystems.
We provide additional examples in Appendix~\ref{sec:example_tools}.

Finally, environment generation, including the environment audit step (Section~\ref{sec:quality-control-filtering}), remains efficient and accessible to labs of all sizes; we report the number of input and output tokens required per environment in Appendix~\ref{app:generation_details}.

\begin{figure}
    \centering
    \begin{subfigure}[b]{0.49\textwidth}
        \centering
        \includegraphics[width=\linewidth]{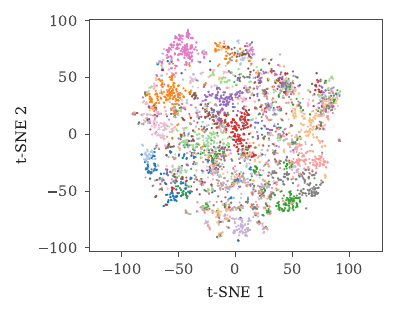}
        \label{fig:tools_embedding_diverse}
    \end{subfigure}
    \hfill
    \begin{subfigure}[b]{0.49\textwidth}
        \centering
        \includegraphics[width=\linewidth]{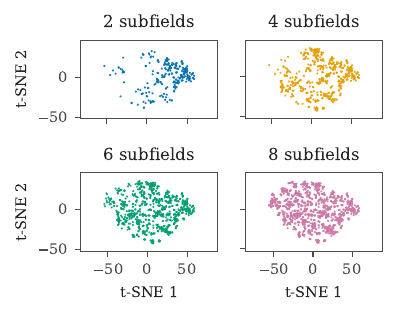}
        \label{fig:tools_embedding_progressive}
    \end{subfigure}
    \caption{\textbf{Distribution of tool embeddings across diverse fields.} \textit{Left:} t-SNE projection of tools spanning 20 fields. \textit{Right:} As we scale the number of tools within a single field (E-commerce and Retail), they grow more diverse rather than producing duplicates.}
    \label{fig:tools_embedding_dist}
    \end{figure}

\section{Environment Simulation and Validation} \label{sec:tool-simulation}
\begin{figure*}
    \centering
    \includegraphics[width=\linewidth]{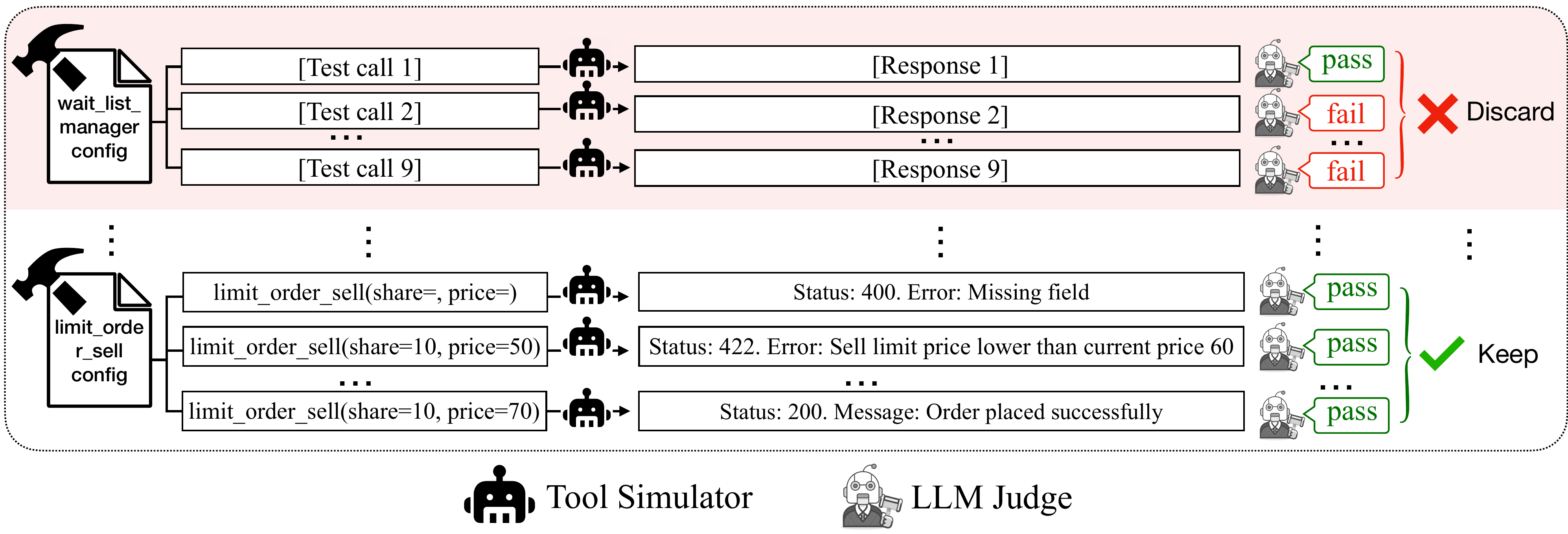}
    \caption{
        Our Tool Validation component uses an LLM judge to filter the tools specifications that are prone to errors. 
    }
    \label{fig:pipeline}
\end{figure*}

This module ensures the generated tools from the previous module can be reliably executed.
\subsection{Tool Simulation}
\label{sec:tool-simulation}
The first step to ensure reliability is to systematically emulate the API call behavior for each tool configuration (e.g. Figure~\ref{fig:example-tool}). 
The key requirement for the tool simulator is  \textit{reliability}. 
A reliable simulator must return appropriate error messages for incorrect or incomplete calls; and generate valid responses for correct ones conditioned on the \textit{metadata} that encodes the current task state. For instance, a reliable flight booking tool should either return \texttt{``no seat available''} or issue a ticket ID depending on whether any seats matching the query (economy, business) remain available (the metadata). Achieving this behavior consistently in practice can be challenging.
To achieve reliability, we decompose the simulation procedure into two distinct stages: \textbf{parameter validation} and \textbf{response generation}. In both stages, we prompt an LLM with access to the tool configuration to emulate tool behavior. Specifically
\begin{enumerate}[leftmargin=*,itemsep=2pt,topsep=2pt,parsep=0pt]
    \item \textbf{Parameter validation.}
    Schema-level checks, tool name, required parameters, and type correctness, are performed programmatically via Abstract Syntax Tree (AST) parsing. The LLM then validates semantic constraints such as cross-field consistency and compatibility with the metadata. If any condition fails, the simulator returns an error message identifying the first issue encountered, along with the corresponding HTTP status code.
    \item \textbf{Response generation for valid calls.} On successful parameter validation, the simulator advances to response generation.
    A key design choice is to incorporate tool-call \textit{metadata} at this stage.
    Depending on the relation between metadata and tool call, the stage proceeds via either \textbf{data generation} or \textbf{information deduction}.
    For valid tool calls unrelated to the metadata, the simulator produces realistic outputs adhering to predefined schemas and domain-specific patterns.
    When the tool call demands reasoning over metadata, the simulator cross-references it to infer the current system state and derive the precise response the API would produce.

\end{enumerate}
To ensure reliable simulator behavior, we refined its prompts through extensive manual testing, iteratively updating them based on observed failure cases until achieving desired performance.
We then evaluated the finalized prompts using two complementary approaches: (1) \textit{manual inspection} of a moderate number of tool responses, and (2) automated assessment via the \textit{Tool Validation} module (Section~\ref{sec:quality-control-filtering}), which uses a language model as a judge to evaluate thousands of simulated responses.

For both experiments, we systematically examine each generated tool with a comprehensive suite of test calls spanning four distinct modes:

\begin{enumerate}[leftmargin=*,itemsep=2pt,topsep=2pt,parsep=0pt]
\item \textbf{Schema failures (basic parameter validation).} Programmatic level errors, e.g., missing required parameters, incorrect parameter types, or malformed inputs that prevent basic function invocation.
    \item \textbf{Constraint failures (tool-specific validation errors).} 
    Mismatched array lengths, invalid value ranges, contradicting parameters, or contradictions with the provided metadata, such as referencing non-existent user records or attempting operations on unavailable resources.
    \item \textbf{Successful executions with new input.} Valid tool call that does not match with any information provided in the metadata, and hence the tool must generate a new reasonable response. 
    \item \textbf{Successful executions with known input.} Valid tool call given as an example in the provided metadata. Note that in this case, since we know the desired output exactly, whether the tool responses match the information given in the metadata can be \textit{automatically} verified.
\end{enumerate}
Refer to Appendix~\ref{sec:failure_success_model} for examples.

\subsubsection{Manual Evaluation}
We evaluate the tool simulator on two tool sources:
(a) tools generated by SynthTools, and
(b) tools from ACEBench~\citep{chen2025acebenchwinsmatchpoint}, which provides deterministic, programmatically defined behaviors. 
Our tool simulator demonstrates high accuracy ($\geq$ \textbf{94\%}) in both cases.
For (a), we evaluate 698 tool responses generated under four modes. The Tool Simulation module achieves an overall accuracy of \textbf{97\%}. Specifically, 200 responses from the first three modes were manually verified, achieving \textbf{94\%} accuracy, while the remaining 498 responses from the fourth mode were automatically evaluated and achieved \textbf{98\%} accuracy.
For source (b), we aligned our simulator's metadata to ACEBench specifications, ensuring consistent initial states. 
Then, using the exact same prompt, we generated approximately eight test calls per tool across 20 ACEBench tools (161 calls overall), encompassing constraint failures and both successful executions modes. 
The simulator matched the ground truth in 151 out of 161 cases, yielding a \textbf{94\%} accuracy, with 14 out of 20 tools achieving perfect agreement (Figure~\ref{fig:ace-examples}). The remaining mismatches primarily arise from implementation-specific ordering differences (e.g., authentication check before or after parameter validation) rather than errors in core simulation logic. 

\begin{figure}
    \centering
    \begin{minipage}[t]{0.49\textwidth}
        \centering
        \begin{tcolorbox}[
          colback=orangePPT!5!white,
          colframe=orangePPT,   
          equal height group=acex,
          title= ACEBench test example
        ]
        \small
        \textbf{Tool call message:}
        \textit{ModifyFlight}(user\_id = 'user1', reservation\_id = 'res\_1', new\_cabin = 'Business Class')

        \textbf{Response:}
        \textit{Status}: PASS, ~
        \textit{Status Code}: 200, ~
        \textit{Return Data}: Cabin upgraded to Business Class. Price difference of 1800 yuan has been charged. Modification has been completed successfully.

        \textbf{Sandbox execution message:}
        Cabin change successful. Price difference paid: 1800.
        \end{tcolorbox}
    \end{minipage}
    \hfill
    \begin{minipage}[t]{0.49\textwidth}
        \centering
        \begin{tcolorbox}[
          colback=orangePPT!5!white,
          colframe=orangePPT,   
          equal height group=acex,
          title= Judge success example
        ]
        \small
        \textbf{Tool call message:}
        \textit{MetricsCalculator}(start\_date = '2024-01-31T23:59:59Z', end\_date = '2024-01-01T00:00:00Z')

        \textbf{Response:}
        \textit{Status}: PASS,  ~
        \textit{Return Data}: ...

        \textbf{Judge Reasoning:}
        Incorrectly specified start\_date='2024-01-31T23:59:59Z' larger than end\_date='2024-01-01T00:00:00Z'. Response should be FAIL with this error message.
        \end{tcolorbox}
    \end{minipage}
    \caption{\textit{Left}: An example ACEBench tool call with simulator return data precisely matching the execution output. \textit{Right:} An example where the judge correctly identifies an erroneous response.}
    \label{fig:ace-examples}

\end{figure}

\subsubsection{LLM-as-Judge Evaluation}
We further evaluate the simulator at scale using an LLM-based judge implemented in the \textit{Tool Validation} module (Section~\ref{sec:quality-control-filtering}).
In this experiment, we focus on the first three modes, whose correctness cannot be assessed automatically. 
We generated 8-10 tool responses for each tool from these modes, and found that \textbf{86\%} were correct and more than \textbf{89\%} had no more than one error across 8–10 stress test calls (see Table \ref{tab:incorrect_responses}), highlighting the high reliability of the generated tools. Interestingly, the audit procedure did not impact the overall complexity of the dataset, leaving the distribution over argument types and number of arguments per tool unchanged (Table \ref{tab:audit_before_after}). 

\begin{table}
\centering
\small
\begin{tabular}{ l c c c c c c c }
\toprule
 & \textbf{\# args (mean)} & \textbf{string} & \textbf{array} & \textbf{float} & \textbf{integer} & \textbf{boolean} & \textbf{object} \\
\midrule
\textbf{Before} & 5.02 & 52.4\% & 15.6\% & 9.8\%  & 10.1\% & 9.3\% & 2.8\% \\
\textbf{After}  & 5.13 & 51.2\% & 16.4\% & 10.5\% & 10.2\% & 9.1\% & 2.7\% \\
\bottomrule
\end{tabular}
\caption{Tool argument statistics before and after audit filtering ($\leq 1$ mistake).}
\label{tab:audit_before_after}
\end{table}

\subsection{Tool Validation}
\label{sec:quality-control-filtering}
As discussed earlier, the \textit{Tool Validation} module supports large-scale evaluation (Section~\ref{sec:tool-simulation}), prunes ill-specified or hard-to-simulate tools (Figure~\ref{fig:pipeline}), and serves as a verifier for on-the-fly checking and retries. For these purposes, it must achieve high accuracy with a low false positive rate.

To achieve this, we employ a carefully engineered \emph{LLM judge}. 
The judge receives a tool specification, test call, and simulator response as input, then returns a structured judgment comprising correct/incorrect status, confidence score, and detailed rationale.
Leveraging the generator–verifier gap~\citep{song2025mind}, we refine the judge using hand-crafted edge cases (e.g., subtle type violations, cross-field inconsistencies) until it reliably detects each error type.

We then evaluate the final judge using stress tests built from deliberately mismatched (tool call, response) pairs. These tests measure both overall \textit{accuracy} and \textit{false positive rate} by assessing whether the judge correctly flags inconsistencies and accepts valid responses.

We group the test cases into three failure modes and three success modes:
(1) \textbf{Failure mode 1:} Incorrect call due to schema violations; simulator mishandles.
(2) \textbf{Failure mode 2:} Schema-valid call that violates documented constraints; simulator mishandles.
(3) \textbf{Failure mode 3:} Fully correct call; simulator output is incorrect (e.g., wrong status code or inconsistent response).
(4) \textbf{Success mode 1:} Incorrect call; simulator handles correctly.
(5) \textbf{Success mode 2:} Correct call that triggers documented constraints; simulator handles correctly.
(6) \textbf{Success mode 3:} Fully correct call; simulator responds correctly. We generated 2-3 (tool call, response) pairs for each failure mode, 1 pair for each success mode, and manually verified the results. In addition, we evaluate judge reliability using 498 mode-4 tool responses (Section~\ref{sec:tool-simulation}), whose correctness is automatically known.

Across 200 manually inspected and 498 automatically graded cases, the LLM judge is correct in 680 cases, achieving \textbf{97\%} accuracy with a \textbf{0\%} false positive rate. On the 200 stress-tests, the judge makes only $3$ errors, and only $6$ actual errors on the automatically graded set, with the remaining $9$ being overly strict judgments rather than genuine misclassifications. These results demonstrate the robustness of the judge and validate its role in the tool validation pipeline.

\section{Bottom-up Task and Trajectory Generation}
\label{sec:dataset}
Finally, complementing the top-down construction of tools, this module re-builds new tasks back up from the validated toolset. Hierarchical generation creates seed tasks to produce coherent, interrelated tools; after filtering removes unreliable ones, we generate new tasks conditioned on the remaining validated tools. Because the tools originate from the same seed task, they retain meaningful dependencies. This makes them particularly well-suited for a sequential downstream workflow. Our composable re-generation pipeline, partially inspired by AgentSynth \cite{xie2025agentsynthscalabletaskgeneration} in the context of computer-use, scales to produce tasks that are \textbf{verifiable}, \textbf{solvable} and \textbf{tunable}.

\begin{figure}
    \centering
    \includegraphics[width=0.85\linewidth]{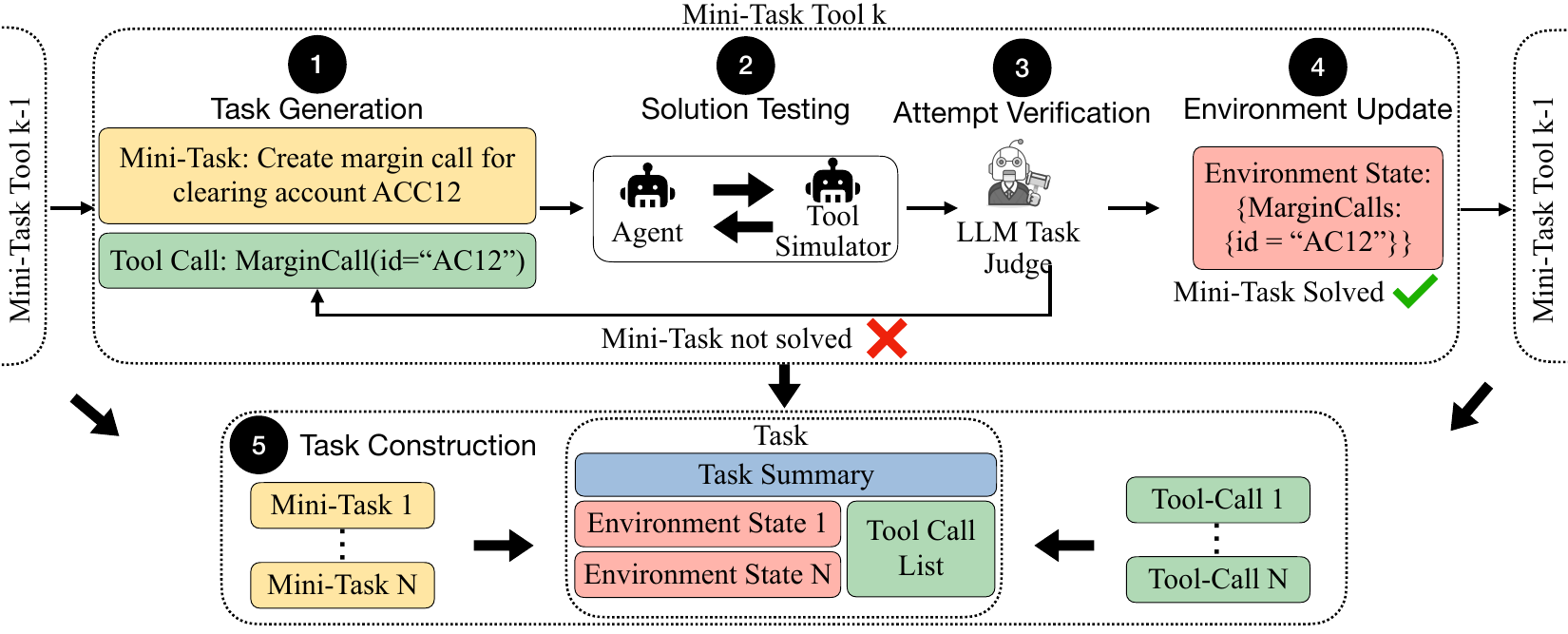}
    \caption{Task Generation Scheme}
    \label{fig:pipeline}
\end{figure}

To achieve this, we leverage the bottom-up idea of starting with atomic capabilities defined by a single tool \citep{lu2026youtullmunlockingnativeagentic}. A similar idea has been explored in deep-research setting with knowledge graph \citep{tongyideepresearchteam2025tongyideepresearchtechnicalreport}.
\begin{enumerate}[leftmargin=*,itemsep=2pt,topsep=2pt,parsep=0pt]
    \item \textbf{Tool-chain proposal.}
    For each validated toolset environment derived from a seed task (typically 10-14 tools), we prompt an LLM to propose multiple \emph{tool-use chains}. Each chain represents a plausible sequential workflow and serves as the backbone of a candidate task. Although not purely combinatorial, the number of viable chains grows rapidly with the size and diversity of the toolset.
    \item \textbf{Mini-task generation and state evolution.} For each proposed chain, we iteratively construct a sequence of mini-tasks (see Appendix \ref{box:task_gen}). At each step $t$, given the current environment state, the system: (1) generates a mini-task description and its ground-truth tool call for the $t$-th tool; (2) lets an agent attempt the task via the simulator; (3) verifies correctness by comparing the agent's tool call with the ground truth and solvability of the mini-task by making sure all arguments needed are grounded in the task description or previous interactions, regenerates if not; and (4) on success, updates the environment using the tool's output and proceeds to the next tool in the chain.

    This yields interdependent mini-tasks whose later steps may depend on earlier outputs. Together with the golden ground truth tool calls and environment state evolution, this forms a coherent task with an explicit solution trajectory and automatic verifiability by comparing the final state of the environment and the sequence of tool-calls.
    \item \textbf{Controlling task difficulty.} Finally, we summarize the sequence of mini-tasks into a complex task description. The difficulty of this task can be tuned by: (1) varying the number of mini-tasks, (2) adjusting the trade-off between explicitness and ambiguity, and (3) adding additional irrelevant but plausible tools in the candidate pool.
\end{enumerate}

We follow this pipeline and construct a dataset of $79,925$ tasks, spanning $100$ fields and $6,800$ tool collections. By design, these tasks are \textbf{verifiable} and \textbf{solvable}: each mini-task is paired with a known ground-truth tool call and a final state of the environment, so chaining them yields an explicit solution trajectory, and every mini-task is successfully executed during generation, ensuring feasibility.

Additionally, tasks are inherently \textbf{tunable} because we can calibrate \textbf{difficulty} and structure by orchestrating the arrangements of tool calls. By alternating the sequencing tool dependencies, we explicitly favor complex \textbf{multi-step} workflows. 

This structured bottom-up approach echoes recent frontier work on synthetic task generation and agentic tool use, which relies on structured task construction rather than naive prompting~\citep{tongyideepresearchteam2025tongyideepresearchtechnicalreport,ramrakhya2025scalingsynthetictaskgeneration}. More broadly, these efforts highlight that effective task generation demands explicit structure and quality control: naive prompting often yields shallow or weakly grounded tasks. Addressing this challenge is critical, as the lack of scalable, rich environments requiring integrated tool use and long-horizon planning remains a key bottleneck in existing agent frameworks~\citep{sheng2024verl, andrews2025arescalingagentenvironments, griggs2025skrylv01, spisak2025openenv}.

Figure~\ref{fig:example-task-1} presents an illustrative example pertaining to the field of e-commerce and retail, including the task specification and the minimal tool set required to complete the task.
Additional examples, covering a range of difficulty levels and various fields are provided in Appendix \ref{sec:example_tasks}.

\begin{figure}
\begin{tcolorbox}[colback=orangePPT!5!white,
                  colframe=orangePPT,
                  title= Task: Defective-Screen Return and Refund,
                  fonttitle=\bfseries,
                  enhanced,
                  segmentation style={draw=none},
                  top=2pt, bottom=2pt, left=4pt, right=4pt,
                  boxsep=2pt
                  ]
{\small\raggedright
\textbf{Task Description:} Process return RET-0001 for customer Alice Johnson (CUST678): 2 units of SKU-ABC-001 from order ORD12345 with a defective screen, exchange requested. Verify eligibility under policy POLICY-2023-RETURN given purchase 2023-08-15T10:30:00Z and check date 2023-09-01T12:00:00Z, generate a prepaid UPS Ground label to warehouse WH-001 at 123 Main St, Anytown, USA 12345, mark RET-0001 Approved and email Alice (alice.johnson@example.com) using template \texttt{return\_approved\_template}, restock the 2 units at WH-001, compute the refund at \$299.99 per unit with 8.25\% tax (no shipping, USD) for \$649.48, process it through Stripe against transaction txn-ORD12345-001, and post the entry under accounting code 4000-REFUND.\\
\textbf{Ground-Truth Tool Sequence:} {ReturnRequestIntake},
{ReturnEligibilityChecker},
{ReturnStatusTracker},
{CustomerNotificationSender},
{InventoryRestockTrigger},
{RefundAmountCalculator},
{RefundProcessor},
{FinancialReconciliationRecorder}\\
\textbf{Initial Environment State:} \texttt{inventory[WH-001][SKU-ABC-001]: \{on\_hand: 0, status: "not\_recorded"\}} \\
\textbf{Final Environment State:} \texttt{inventory[WH-001][SKU-ABC-001]: \{on\_hand: 2, restock\_id: "RST-20230901-001"\}}
\par}
\end{tcolorbox}
\caption{An example of a verifiable task generated from our pipeline for the field \texttt{E-Commerce and Retail} together with its ground truth sequence of tool calls and state of the environment.}
\label{fig:example-task-1}
\end{figure}

\section{Case Study: Midtraining with SynthTools}\label{sec:midtraining}
\label{sec:case_study}
To demonstrate a concrete application of \textit{SynthTools}, we run a midtraining case study. Starting from Qwen3 Base models (size 0.6B, 1.7B \& 4B), we continue training on trajectories generated from the previous task dataset, and evaluate the resulting models on multiple agentic tool-use benchmarks. Across benchmarks, midtrained models show consistent performance gains, demonstrating that synthetic tool interactions help improve tool-use capabilities.
\subsection{Midtraining}
Recent work highlights the value of a midtraining phase, a continued pretraining stage appended after large-scale pretraining. Although like pretraining but unlike post-training, this phase typically still trains on next-token prediction objective, except with higher quality, often instruction-formatted dataset. This stage aims to reduce the distribution gap between generic pretraining text and downstream high-level tasks requiring reasoning, tool use, and structured interaction \cite{ tongyideepresearchteam2025tongyideepresearchtechnicalreport, lu2026youtullmunlockingnativeagentic}. Effective midtraining corpora must therefore be both scalable and structurally diverse. SynthTools is designed precisely for this role: it generates large-scale, verifiable, and diverse tool-use trajectories. This enables continued pretraining on interaction-heavy data and promoting tool-use capabilities.

\subsection{Experiments and Results}
We midtrain Qwen3-0.6B, 1.7B, and 4B Base models on standard next-token prediction loss over trajectories generated by a GPT-OSS-120B \cite{openai2025gptoss120bgptoss20bmodel} agent interacting with our task dataset. We evaluate the midtrained (MT) models on 3 benchmarks: API-Bank~\citep{li_apibank}, ACEBench~\citep{chen2025acebenchwinsmatchpoint}, and the Berkeley Function Calling Leaderboard (BFCL)~\citep{patil2025bfcl}. We select benchmark categories measuring tool/API use rather than instruction following or chat. Notably, API-Bank evaluates $73$ realistic, executable implementations of commonly used real-world APIs, and BFCL-v3 Live evaluates against real functioning APIs, directly demonstrating synthetic-to-real transfer. Improvements on these benchmarks therefore suggest that capabilities learned from synthetic tools can transfer to realistic API-style interactions.

Results are summarized in Table~\ref{tab:midtrain}. Across model sizes and task categories, midtrained models consistently outperform their corresponding base models, demonstrating that midtraining on SynthTools generally improves agentic tool-use ability.

\begin{table*}
\centering
\footnotesize
\setlength{\tabcolsep}{6pt}
\renewcommand{\arraystretch}{1.15}
\begin{tabular}{llcccc}
\toprule
\textbf{Size} & \textbf{Type} & \textbf{ACEBench (Normal)} & \textbf{BFCL (Non Live)} & \textbf{BFCL (Live)} & \textbf{API-Bank (L1)} \\
\midrule
\multirow{2}{*}{0.6B} & base              & 0.307 & 0.297 & 0.098 & 0.591 \\
                      & midtrained (ours) & \textbf{0.397} & \textbf{0.333} & \textbf{0.205} & \textbf{0.664} \\
\midrule
\multirow{2}{*}{1.7B} & base              & 0.377 & 0.628 & 0.393 & 0.611 \\
                      & midtrained (ours) & \textbf{0.466} & \textbf{0.655} & \textbf{0.409} & \textbf{0.686} \\
\midrule
\multirow{2}{*}{4B}   & base              & 0.583 & 0.690 & 0.432 & 0.761 \\
                      & midtrained (ours) & \textbf{0.651} & \textbf{0.701} & \textbf{0.538} & \textbf{0.813} \\
\bottomrule
\end{tabular}
\caption{Results of Base and midtrained Qwen3 models on ACEBench~\cite{chen2025acebenchwinsmatchpoint}, BFCL~\cite{patil2025bfcl}, and API-Bank~\cite{li_apibank}. Midtraining on SynthTools improves performance across all model sizes and benchmarks.}
\label{tab:midtrain}
\end{table*}

\section{Conclusion} \label{sec:Conclusion}

We introduce \textit{SynthTools}, a scalable and structured framework for building synthetic tool ecosystems and verifiable tasks for agent training and evaluation. Our results show that such environments serve as an effective dataset for continued pretraining, pointing toward synthetic interaction corpora as a key ingredient for advancing tool-using AI systems.

\begin{ack}
This research used resources of the Oak Ridge Leadership Computing Facility (OLCF) and Argonne Leadership Computing Facility (ALCF) which are a DOE Office of Science User Facilities. This work was supported by an award from the ASCR Leadership Computing Challenge (ALCC) under project ERCAP0034861.
\end{ack}

\bibliography{ref}
\clearpage

\appendix

\section{Validation Details}
\label{app:audit-details}

In this section we provide additional details about the audit step described in Section~\ref{sec:quality-control-filtering}. We report tool-level reliability across stress tests, per-mode accuracy of the simulator, and a breakdown of the verdict types associated with the simulator's mistakes.

\begin{table}[H]
\centering
\small
\begin{tabular}{ c c c c c }
\toprule
\text{Number of Incorrect Responses} & \textbf{0} & \textbf{1} & \textbf{2} & \textbf{3} \\
\midrule
\text{Percentage of Tools} & 57\% & 32\% & 7\% & 3\% \\
\addlinespace[0.2em]
\specialrule{1pt}{0pt}{0pt}
\addlinespace[0.2em]
\textbf{Overall Accuracy} & \multicolumn{4}{c}{\textbf{94\%}} \\
\bottomrule
\end{tabular}
\caption{Distribution of incorrect responses among $8$--$10$ stress test calls and overall tool response accuracy as judged by the \textit{Tool Validation} module.}
\label{tab:incorrect_responses}
\end{table}

At the tool level (Table~\ref{tab:incorrect_responses}), a majority of tools ($57\%$) produce no incorrect responses, and $89\%$ have at most one. Only $10\%$ exhibit two or more failures, indicating that simulation errors are concentrated in a small tail of hard-to-simulate tools. At the response level, the simulator achieves $94\%$ overall accuracy.

\begin{table}[H]
    \centering
    \small
    \begin{tabular}{ l c }
    \toprule
    \textbf{Mode} & \textbf{\% Correct} \\
    \midrule
    \text{Failure Mode 1} & 95.9\% \\
    \text{Failure Mode 2} & 83.6\% \\
    \text{Failure Mode 3} & 82.5\% \\
    \addlinespace[0.2em]
    \specialrule{1pt}{0pt}{0pt}
    \addlinespace[0.2em]
    \textbf{Overall Accuracy} & \textbf{86.2\%} \\
    \bottomrule
    \end{tabular}
    \caption{Per-mode accuracy across Failure Mode 1 (schema/type), Failure Mode 2 (semantic rule), and Failure Mode 3 (happy path) checks, with overall accuracy aggregated across all responses.}
    \label{tab:mode_accuracy}
\end{table}

Broken down by test mode (Table~\ref{tab:mode_accuracy}), schema and type checks (Mode 1) are handled almost perfectly at $95.9\%$, since they are largely delegated to programmatic AST parsing. Performance drops on Mode 2 ($83.6\%$), which requires reasoning about semantic constraints, and is lowest on Mode 3 ($82.5\%$). The dominant source of error in Mode 3 is \emph{underspecified tool schemas}: when a schema does not document how a particular success or error case should be handled, the simulator is forced to hallucinate the missing behavior. These cases are addressed by dropping the affected tools in the filtering step.

\begin{table}[H]
\centering
\small
\begin{tabular}{ l c }
\toprule
\textbf{Verdict} & \textbf{\% of Mistakes} \\
\midrule
\texttt{should\_success\_got\_error} (sim invented an error on a valid call) & 58.3\% \\
\texttt{inappropriate\_response} (right status, wrong/malformed payload)     & 31.5\% \\
\texttt{should\_error\_got\_success} (sim missed a violation, returned 200)  & 10.0\% \\
\bottomrule
\end{tabular}
\caption{Distribution of failure modes among incorrect simulator responses, categorized by verdict type.}
\label{tab:mistake_verdicts}
\end{table}

Categorized by verdict type (Table~\ref{tab:mistake_verdicts}), the dominant failure is \texttt{should\_success\_got\_error} ($58.3\%$), where the simulator over-rejects a valid call. Another $31.5\%$ are \texttt{inappropriate\_response} cases, with correct status codes but malformed payloads. Only $10\%$ are \texttt{should\_error\_got\_success}, where the simulator misses a genuine violation. This skew is favorable: false rejections trigger regenerations via the retry mechanism, whereas silent false acceptances would propagate noise into the dataset.

\subsection{Failure and success models of tool calls}
\label{sec:failure_success_model}

\begin{minipage}[b]{0.49\textwidth}
        \centering
\begin{tcolorbox} [
  colback=orangePPT!5!white,
colframe=orangePPT,   
  height=5.4cm,      
  title= Failure mode 1: Schema mismatch
]
\small
\textbf{Tool call message:}

\textit{Insurance\_Information\_Updater}~ (patient\_id = 'PAT001', insurance\_fields = [], insurance\_values = ['Blue Cross'])

\medskip
\textbf{Response:}

\textit{Status}: FAIL, ~
  \textit{Status Code}: 400, ~
     \textit{Error Message}: Invalid parameter: insurance\_fields array must contain at least 1 item."
     
\end{tcolorbox}
    \end{minipage}
\hfill
\begin{minipage}[b]{0.49\textwidth}
        \centering
\begin{tcolorbox}  [
  colback=orangePPT!5!white,
colframe=orangePPT,   
  height=5.4cm,      
    title= Failure mode 2: Parameter inconsistency
]
\small
\textbf{Tool call message:}

\textit{Insurance\_Information\_Updater}~(patient\_id = 'PAT001', insurance\_fields = ['provider', 'policy\_number'], insurance\_values = ['Blue Cross'])

\medskip
\textbf{Response:}
 
 \textit{Status}: FAIL,
  \textit{Status Code}: 400, 
 \textit{Error Message}: Mismatched fields and values: Ensure insurance\_fields and insurance\_values arrays have the same length.

\end{tcolorbox}
    \end{minipage}
\hfill
\begin{minipage}[b]{0.49\textwidth}
        \centering
\begin{tcolorbox}  [
  colback=orangePPT!5!white,
colframe=orangePPT,   
  height=5cm,      
  title= Failure mode 2: Metadata inconsistency
]
\small
\textbf{Tool call message:}\\
\textit{Regulation\_Detail\_Fetcher}~(regulation\_id = 'INVALID-REG-999')

\medskip
\textbf{Response:}

\textit{Status}: PASS, 
 \textit{Status Code}: 200, 
\textit{Return Data}: Error: Invalid regulation ID: Ensure the regulation ID is correct and exists in the database.

\end{tcolorbox}
    \end{minipage}
    \hfill
\begin{minipage}[b]{0.49\textwidth}
        \centering
\begin{tcolorbox}  [
  colback=orangePPT!5!white,
colframe=orangePPT,   
  height=5cm,      
  title = Success mode: Correct response
]
\small
\textbf{Tool call message:}

\textit{Insurance\_Information\_Updater} ~ (patient\_id = 'PAT001', insurance\_fields = ['provider'], insurance\_values = ['Blue Cross'])

\medskip
\textbf{Response:}

 \textit{Status}: PASS, 
   \textit{Status Code}: 200, 
    \textit{Return Data}: update\_status: Success,  updated\_insurance: ['provider']

\end{tcolorbox}
    \end{minipage}

\section{Midtraining with SynthTools}
Training was performed on a corpus of ${\sim}23{,}000$ trajectories for a total of $200$M tokens using the following hyperparameters on $4\times$NVIDIA A100 GPUs 40\,GB. Evaluation was performed at temperature $T = 0$ (greedy decoding); consequently, no confidence intervals are reported, as the evaluation is deterministic.
\begin{table}[H][h]
\centering
\small
\begin{tabular}{ l c c c }
\toprule
\textbf{Hyperparameter} & \textbf{Qwen3-0.6B} & \textbf{Qwen3-1.7B} & \textbf{Qwen3-4B} \\
\midrule
Optimizer              & AdamW              & AdamW              & AdamW              \\
Adam $\beta_1$         & 0.9                & 0.9                & 0.9                \\
Adam $\beta_2$         & 0.999              & 0.999              & 0.999              \\
Adam $\epsilon$        & $1\times10^{-8}$   & $1\times10^{-8}$   & $1\times10^{-8}$   \\
Weight decay           & 0.0                & 0.0                & 0.0                \\
Learning rate          & $2\times10^{-6}$   & $1\times10^{-6}$   & $8\times10^{-7}$   \\
LR schedule            & cosine             & cosine             & cosine             \\
Warmup ratio           & 0.05               & 0.05               & 0.05               \\
Epochs                 & 3                  & 3                  & 3                  \\
Training steps         & 1{,}976            & 1{,}976            & 1{,}976            \\
Max sequence length    & 16{,}384           & 16{,}384           & 16{,}384           \\
Per-device batch size  & 2                  & 2                  & 2                  \\
Gradient accumulation  & 4                  & 4                  & 2                  \\
Max gradient norm      & 1.0                & 1.0                & 1.0                \\
Random seed            & 23                 & 23                 & 23                 \\
\bottomrule
\end{tabular}
\caption{Training hyperparameters for the Qwen3 models. Adam $\beta_1$, $\beta_2$, $\epsilon$ and the weight decay are the Hugging Face \texttt{Trainer} defaults.}
\end{table}

\section{Pipeline Details}

\subsection{Computational Detail} \label{app:generation_details}
All generation was performed using \texttt{gpt-oss-120b}\cite{openai2025gptoss120bgptoss20bmodel} on $4\times$NVIDIA A100 GPUs 40\,GB.

\begin{table}[H][h]
\centering
\small
\begin{tabular}{ l r r r }
\toprule
\textbf{Stage} & \textbf{Input} & \textbf{Output} & \textbf{Total} \\
\midrule
\text{1. Environment generation}       & 11{,}732  & 21{,}533  & 33{,}265  \\
\quad \text{Subfields + tasks + tools} & 5{,}013   & 9{,}108   & 14{,}121  \\
\quad \text{Sequences build}           & 6{,}720   & 12{,}425  & 19{,}144  \\
\text{2. Environment audit}            & 748{,}459 & 194{,}880 & 943{,}338 \\
\addlinespace[0.2em]
\specialrule{1pt}{0pt}{0pt}
\addlinespace[0.2em]
\textbf{Total per environment} & \textbf{760{,}191} & \textbf{216{,}412} & \textbf{976{,}603} \\
\addlinespace[0.2em]
\specialrule{1pt}{0pt}{0pt}
\addlinespace[0.2em]
\text{Task generation} & 190{,}308 & 69{,}303 & 259{,}611 \\
\bottomrule
\end{tabular}
\caption{Token usage per environment and per task across pipeline stages.}
\label{tab:token-cost}
\end{table}

\section{Deduplication details}
\label{sec:deduplication_details}

To eventually eliminate near-duplicate tools generated through hierarchical evolution, we employ a multi-stage de-duplication pipeline.  The process begins with exact de-duplication applied independently within each domain/field.  We first normalize and compare the \texttt{tool\_name} attributes, followed by the \texttt{tool\_body}, which encompasses the description, parameters, usage, and output schema.  This step reduces redundancy and prevents unnecessary computation in subsequent stages.

We then implement a semantic de-duplication pipeline, adapted from SemDeDup, a state-of-the-art method proposed by \cite{abbas2023semdedupdataefficientlearningwebscale}, tailored to the structural characteristics of our dataset. The semantic phase proceeds as follows:

We construct an embedding-based similarity graph over the normalized  \texttt{tool\_name} and \texttt{tool\_body} fields. Specifically, let $e(t)\in\mathbb{R}^d$ denote the embedding of tool $t$, and define the cosine similarity between tool $t_i$ and $t_j$ as  $S_{ij}=\cos\!\big(e(t_i),e(t_j)\big)$. An adjacency matrix is then defined as 
$
A_{ij}={\mathbf{1}}\{S_{ij}\ge\tau\},$   $ \tau\in(0,1),
$
and the corresponding undirected similarity graph is
$
\mathcal{G}=\big\{(i,j):A_{ij}=1\big\}
$.  Connected components in $\mathcal{G}$ 
 represent candidate duplicate sets: singleton nodes are considered unique tools, whereas multi-node components are treated as clusters of near-duplicates. We then apply a selection algorithm (Algorithm~\ref{alg:selection-rule}) to retain representative tools from each component.

\begin{figure}[H]
    \centering
    \includegraphics[width=0.6\linewidth]{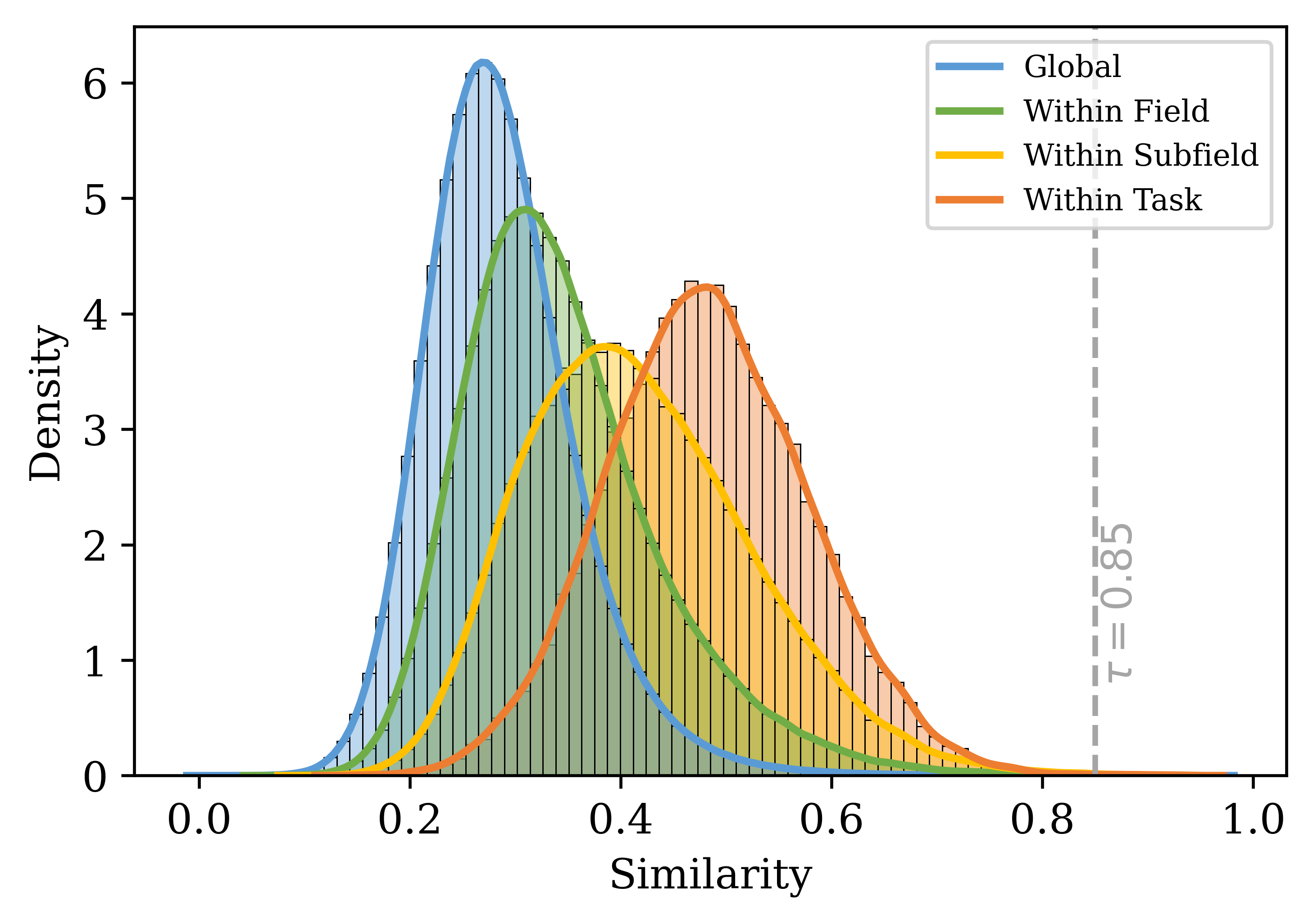}
    \caption{Distribution of semantic similarity scores between tools at varying levels of domain granularity: across fields (blue), within a field (green), within a subfield (yellow), and within a task (red).}
    \label{fig:hist1}
\end{figure}

\begin{figure}[H]
    \centering
    \includegraphics[width=0.6\linewidth]{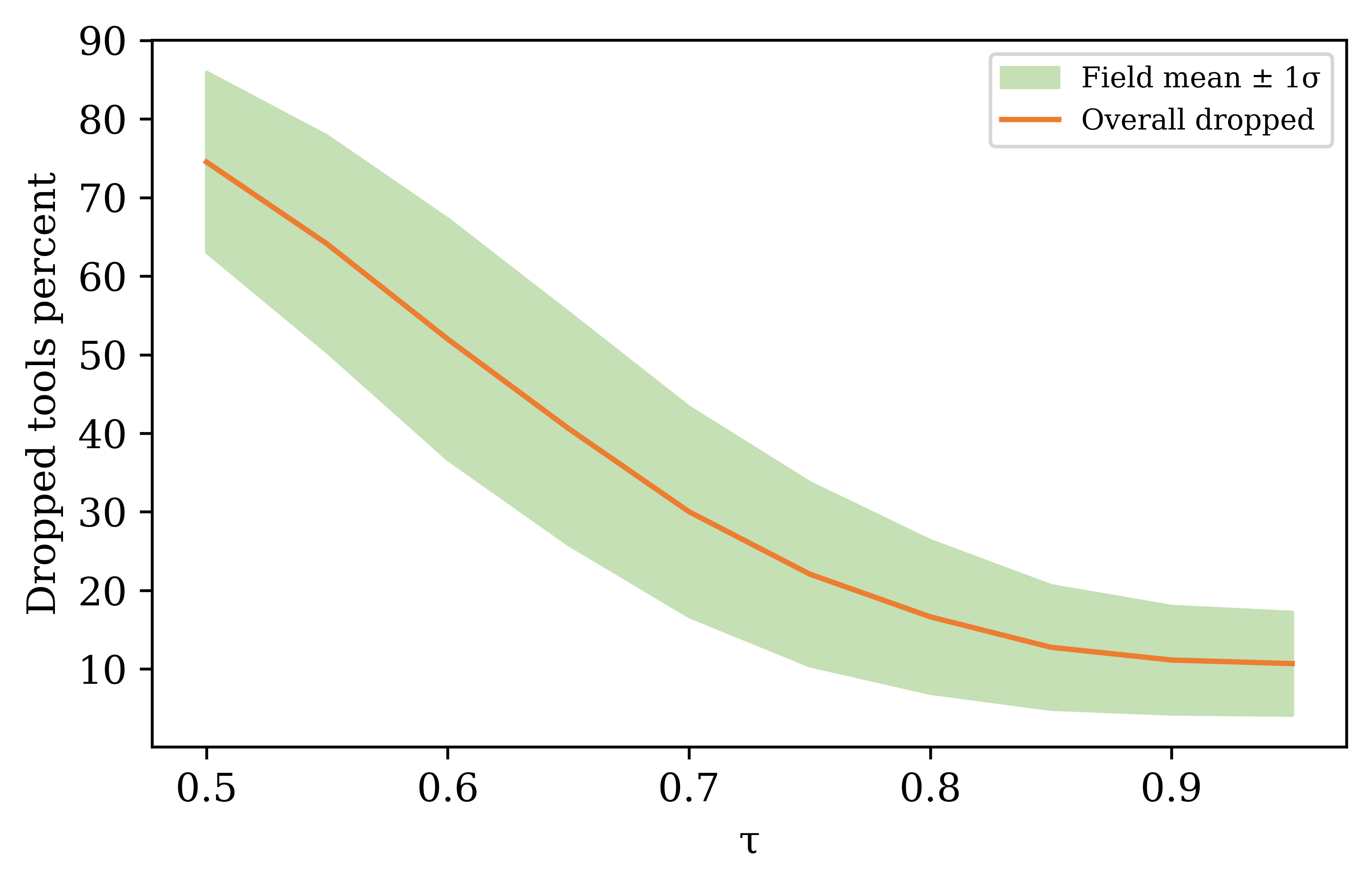}   
    \caption{ Percentage of tools dropped v/s similarity score threshold } 
    \label{fig:thre1}
\end{figure}

\begin{algorithm}
\caption{Selection rule for connected components}
\label{alg:selection-rule}
\begin{algorithmic}[1]
\For{each connected component $C \subseteq \{1,\dots,|T|\}$}
  \If{$|C| = 1$} retain the tool unchanged.
  \ElsIf{$|C| = 2$}  choose one uniformly at random (with a fixed PRNG seed for reproducibility), discard the other.
  \Else \Comment{$|C| \ge 3$}
    \State Initialize $L \gets C$.
    \While{$\exists\, i \neq j \in L$ with $u_i^\top u_j \ge \tau$}
      \For{each $i \in L$}
        \State Compute degree: 
        $
          \deg_\tau(i) = \sum_{j \in L \setminus \{i\}} A_{ij}
        $
        \State Compute incident-sum:
        $
          w_\tau(i) = \sum_{j \in L \setminus \{i\}} A_{ij}\,(u_i^\top u_j)
        $
      \EndFor
      \State Select node to drop by lexicographic maximization:
      $
        v^\star \in \arg\max_{i \in L} \big(\deg_\tau(i),\, w_\tau(i)\big)
      $
      \State Let 
      $
        k^\star = \arg\max_{j \in L \setminus \{v^\star\}} u_{v^\star}^\top u_j
      $
      \State Check condition: $u_{v^\star}^\top u_{k^\star} \ge \tau$.
      \State Drop $v^\star$: update $L \gets L \setminus \{v^\star\}$.
    \EndWhile
    \State Return survivors $L$ for component $C$.
  \EndIf
\EndFor
\end{algorithmic}
\end{algorithm}
Empirically, cross-field duplication is rare, as workflows and vocabularies tend to differ significantly across domains (Figure~\ref{fig:hist1}). Most residual redundancy occurs between adjacent tasks and subdomains, where tool functionalities partially overlap. The degree of de-duplication depends on the threshold $\tau$; Figure~\ref{fig:thre1} illustrates the elimination rate as a function of $\tau$. Based on a validation sweep that balances compactness and coverage, we set $\tau = 0.85$ in practice.

\section{Tool Examples}
\label{sec:example_tools}
In this section, we show a few examples of the tools generated as described in Section~\ref{sec:tool-generation}. 
\subsection{Example 1 (Field: Financial Trading) - Given in Main Paper}
\begin{figure}[H]
{ \small
 \begin{tcolorbox}[colback=orangePPT!5!white,
                  colframe=orangePPT,
                  title=Field to Tool Evolution,
                  fonttitle=\bfseries]
\textbf{Field:} Financial Trading $\rightarrow$ \textbf{Sub Domain}:  Options Trading Strategies $\rightarrow$ \textbf{Task:} Calculate implied volatility for specific option contracts 
$\rightarrow$ \textbf{Tool:}  Black Scholes Calculator  
\end{tcolorbox}
\begin{tcolorbox}[colback=orangePPT!5!white,
                  colframe=orangePPT,
                  title= The Generated Tool: Black Scholes Calculator,
                  fonttitle=\bfseries]


\textbf{Description:}  
Calculates option prices using the Black–Scholes–Merton model. Can price European call and put options based on inputs like underlying price, strike, time to expiration, risk-free rate, and volatility.  

\vspace{1mm}
\textbf{Parameters:}
\begin{itemize}[leftmargin=*]
  \item \textbf{option\_type} (string, required) – Type: \texttt{call} or \texttt{put}
  \item \textbf{underlying\_price} (number, required) – Current price of the underlying asset (\(S\))
  \item \textbf{strike\_price} (number, required) – Strike price of the option (\(K\))
  \item \textbf{time\_to\_expiration} (number, required) – Time to expiration in years (\(T\))
  \item \textbf{risk\_free\_rate} (number, required) – Annual risk-free interest rate in decimal form (\(r\))
  \item \textbf{volatility} (number, required) – Annual volatility in decimal form (\(\sigma\))
  \item \textbf{dividend\_yield} (number, optional, default = \texttt{0}) – Annual dividend yield in decimal form (\(q\))
\end{itemize}
\textbf{Error Messages:}
\begin{itemize}[leftmargin=*]
  \item Invalid option type: \texttt{option\_type} must be either \texttt{call} or \texttt{put}.
  \item Invalid price input: \texttt{underlying\_price} and \texttt{strike\_price} must be positive numbers. 
  \item Invalid time input: \texttt{time\_to\_expiration} must be a positive number (in years).
  \item Invalid rate input: \texttt{risk\_free\_rate} must be a decimal number between 0 and 1. 
  \item Invalid volatility input: \texttt{volatility} must be a positive number.
  \item Invalid dividend yield: \texttt{dividend\_yield} must be a non-negative number. 
\end{itemize}
\textbf{Usage:}  
Provide \texttt{option\_type},
\texttt{underlying\_price}, \texttt{strike\_price}, \texttt{time\_to\_expiration}, \texttt{risk\_free\_rate}, and \texttt{volatility}
 to calculate the theoretical option price. Optionally include \texttt{dividend\_yield} for dividend-paying assets.  

\vspace{1mm}
\textbf{Output Details:}
\begin{itemize}[leftmargin=*]
  \item \textbf{option\_price} (number) – Calculated theoretical price of the option
  \item \textbf{delta} (number) – Option’s delta (sensitivity to changes in underlying price)
  \item \textbf{gamma} (number) – Option’s gamma (sensitivity of delta to changes in underlying price)
  \item \textbf{theta} (number) – Option’s theta (sensitivity to time decay)
  \item \textbf{vega} (number) – Option’s vega (sensitivity to changes in volatility)
  \item \textbf{rho} (number) – Option’s rho (sensitivity to changes in risk-free rate)
\end{itemize}

\end{tcolorbox}

\caption{An example of tool generated through our pipeline for the \texttt{Financial Trading} field.}
\label{fig:example-tool-full}}
\end{figure}

\subsection{Example 2 (Field: Ecommerce and Retail)}

{ \small
 \begin{tcolorbox}[colback=orangePPT!5!white,
                  colframe=orangePPT,
                  title=Field to Tool Evolution,
                  fonttitle=\bfseries]
\textbf{Field:} Ecommerce and Retail $\rightarrow$ \textbf{Sub Domain}: Product Catalog Management $\rightarrow$ \textbf{Task:} Product Quality Assurance and Content Validation 
$\rightarrow$ \textbf{Tool:} Image Quality Analyzer
\end{tcolorbox}

\begin{tcolorbox}[colback=orangePPT!5!white,
                  colframe=orangePPT,
                  title= The Generated Tool,
                  fonttitle=\bfseries,
                  enhanced,
                  breakable,
                  segmentation style={draw=none}]

\textbf{Tool Name:} Image Quality Analyzer  

\textbf{Description:}  
Analyzes product images for quality metrics including resolution, composition, lighting, background, and technical specifications to ensure catalog standards.  

\vspace{2mm}
\textbf{Parameters:}
\begin{itemize}
  \item \textbf{image\_urls} (array of strings, required, 1–20 items) – URLs or file paths of images to analyze
  \item \textbf{min\_resolution} (integer, optional, default = 800) – Minimum required resolution in pixels
  \item \textbf{check\_background} (boolean, optional, default = \texttt{true}) – Whether to analyze background cleanliness
\end{itemize}

\vspace{2mm}
\textbf{Error Messages:}
\begin{itemize}
  \item Image not accessible: One or more image URLs could not be accessed or loaded.
  \item Unsupported image format: Images must be in JPEG, PNG, or WebP format.
  \item Image resolution too low: Image resolution is below the specified minimum requirement.
  \item Image processing failed: Technical error occurred while analyzing image quality.
\end{itemize}

\vspace{2mm}
\textbf{Usage:}  
Provide an array of \texttt{image\_urls} to analyze. Optionally set \texttt{min\_resolution} and background checking preferences. Returns comprehensive quality analysis for each image.  

\vspace{2mm}
\textbf{Output Details:}
\begin{itemize}
  \item \textbf{overall\_score} (number) – Overall image quality score from 0 to 100
  \item \textbf{image\_analyses} (array of strings) – Individual analysis results for each image
  \item \textbf{quality\_issues} (array of strings) – Identified quality problems across all images
  \item \textbf{recommendations} (array of strings) – Suggestions for improving image quality
\end{itemize}

\end{tcolorbox}
}

\section{Example Tasks}
\label{sec:example_tasks}

In this section, we show a few examples of the tasks generated from SynthTools. 

\subsection{Example 1 (Field: Ecommerce and Retail) }
\label{sec:example_1_tasks}
\begin{tcolorbox}[colback=orangePPT!5!white,
                  colframe=orangePPT,
                  title= Task Generation,
                  fonttitle=\bfseries,
                  enhanced,
                  breakable,
                  segmentation style={draw=none},
                  label={box:task_gen}]
{\small
\textbf{Tool Sequence}
\noindent
MetricComparisonSuite $\rightarrow$ 
ComprehensiveReportGenerator $\rightarrow$ 
MetricFormatter $\rightarrow$ 
AlertGenerator $\rightarrow$ 
DashboardUpdater $\rightarrow$ 
Scheduler $\rightarrow$ 
ReportBuilder $\rightarrow$ 
MetricFormatter $\rightarrow$ 
AlertGenerator
\vspace{0.5cm}
\hrule
\vspace{0.5cm}
\textbf{Tool 1: MetricComparisonSuite Execution}
\vspace{2mm}
\textbf{Step 1: Task Generation}
\begin{enumerate}
  \item \textbf{Tool Name:} MetricComparisonSuite
  \item \textbf{Target:} engine\_42 (Recommendation Engine)
  \item \textbf{Timeframe:} Test 2023-09-01 to 2023-09-07; Baseline 2023-08-01 to 2023-08-07
  \item \textbf{Configuration:} 
    \begin{itemize}
        \item \textit{Metrics:} CTR (threshold 0.5, above), Conversion Rate (1.0, above), Revenue Lift (5.0, above)
        \item \textit{Output:} Markdown format, Title: ``Weekly Performance Review'', ASCII bar charts
        \item \textit{Delivery:} Email to alice@example.com, bob@example.com
    \end{itemize}
  \item \textbf{Context:} First step of the chain; no prior environment state.
\end{enumerate}
\vspace{2mm}
\textbf{Step 2: Agent Mini-task Execution}
\noindent \textit{Interaction Log:}
\begin{enumerate}
  \item \textbf{Assistant:} \textit{(Decision)} Calls MetricComparisonSuite on engine\_42, comparing the test window against the baseline across CTR, Conversion Rate, and Revenue Lift at the configured thresholds. Requests markdown output with ASCII bar charts titled ``Weekly Performance Review'' and emails both recipients.
  \item \textbf{Tool:} \textit{(Status 200)} Comparison summary generated successfully. Returns a markdown body with one section per metric, including test vs.\ baseline values and ASCII bar charts.
\end{enumerate}
\vspace{2mm}
\textbf{Step 3: Ground-Truth Validation}
\begin{itemize}
  \item \textbf{Tool Call Equality:} True
  \item \textbf{Task Solved:} True (Confidence: 0.98)
  \item \textbf{Explanation:} The agent's generated call perfectly matches the required ground truth parameters.
\end{itemize}
\vspace{2mm}
\textbf{Step 4: Environment Update}
\noindent \textit{Metadata Snapshot:}
\begin{enumerate}
  \item \textbf{Engine ID:} engine\_42
  \item \textbf{Latest Comparison:} ``Weekly Performance Review'' (markdown), test window 2023-09-01 to 2023-09-07 vs.\ baseline 2023-08-01 to 2023-08-07
\end{enumerate}
}
\end{tcolorbox}

\begin{tcolorbox}[colback=orangePPT!5!white,
                  colframe=orangePPT,
                  title= Task Trajectory,
                  fonttitle=\bfseries,
                  enhanced,
                  breakable,
                  segmentation style={draw=none}]
{\small \raggedright
\textbf{Initial Task.} Generate a metric comparison summary for engine \texttt{engine\_42} between the test period 2023-09-01T00:00:00Z to 2023-09-07T23:59:59Z and the baseline period 2023-08-01T00:00:00Z to 2023-08-07T23:59:59Z. Compare the metrics CTR, Conversion Rate, and Revenue Lift with thresholds 0.5, 1.0, and 5.0 respectively, where an increase (\texttt{above}) is considered good for all. Use the report title \texttt{Weekly Performance Review}, output format markdown, include ASCII bar charts, and email the resulting summary to \texttt{alice@example.com} and \texttt{bob@example.com}.
\vspace{1mm}
\hrule
\vspace{2mm}
\begin{itemize}[leftmargin=1.5em, itemsep=3pt]
\item \textbf{Agent:} \texttt{MetricComparisonSuite(engine\_id="engine\_42", test\_start\_time="2023-09-01T00:00:00Z", \dots, email\_recipients=["alice@example.com", "bob@example.com"])}
\item \textbf{Tool:} \texttt{\{"status\_code": 200, "response": \{"comparison\_summary": "\#\#\# Weekly Performance Review\textbackslash n- CTR: \dots\textbackslash n- Conversion Rate: \dots\textbackslash n- Revenue Lift: \dots"\}\}}
\item \textbf{Agent:} \texttt{AlertGenerator(metric\_name="Revenue Lift", current\_value=18.5, threshold=5.0, direction="above", severity="high")}
\item \textbf{Tool:} \texttt{\{"status\_code": 200, "response": \{"alert\_message": "Alert: Revenue Lift is 18.5, which is above the threshold of 5.0. Severity: high."\}\}}
\item \textbf{Agent:} \texttt{DashboardUpdater(dashboard\_id="dashboard\_42", metric\_name="Revenue Lift", metric\_value=18.5, timestamp="2023-09-07T23:59:59Z")}
\item \textbf{Tool:} \texttt{\{"status\_code": 200, "response": \{"update\_status": "Updated Revenue Lift on dashboard dashboard\_42 at 2023-09-07T23:59:59Z."\}\}}
\item[] \hspace{2em}$$\vdots$$
\item \textbf{Agent:} \texttt{AlertGenerator(metric\_name="CTR", current\_value=2.9, threshold=0.5, direction="above", severity="medium")}
\item \textbf{Tool:} \texttt{\{"status\_code": 200, "response": \{"alert\_message": "Alert: CTR is 2.9, which is above the threshold of 0.5. Severity: medium."\}\}}
\item \textbf{Agent:} \texttt{DashboardUpdater(dashboard\_id="dashboard\_42", metric\_name="CTR", metric\_value=2.9, timestamp="2023-09-07T23:59:59Z")}
\item \textbf{Tool:} \texttt{\{"status\_code": 200, "response": \{"update\_status": "Updated CTR on dashboard dashboard\_42 at 2023-09-07T23:59:59Z."\}\}}
\end{itemize}
}
\end{tcolorbox}

\subsection{Example 2 (Field: Financial Trading)}

\begin{tcolorbox}[colback=orangePPT!5!white,
                  colframe=orangePPT,
                  title= Task Generation,
                  fonttitle=\bfseries,
                  enhanced,
                  breakable,
                  segmentation style={draw=none},
                  label={box:task_gen}]
{\small \raggedright
\textbf{Tool Sequence.} TradeRecordIngestor $\rightarrow$ RecordValidator $\rightarrow$ SchemaNormalizer $\rightarrow$ MetadataTagger $\rightarrow$ RetentionPolicyResolver $\rightarrow$ RecordEncryptor $\rightarrow$ SecureArchiveWriter $\rightarrow$ RetentionComplianceChecker.
\vspace{1mm}
\hrule
\vspace{2mm}
\textbf{Tool 1: TradeRecordIngestor Execution}
\vspace{2mm}

\textbf{Step 1: Mini-task Construction.}
\begin{itemize}[leftmargin=1.5em, itemsep=2pt]
  \item \textbf{Tool Name:} \texttt{TradeRecordIngestor}
  \item \textbf{Target:} \texttt{file\_2023\_09.csv} (CSV trade-record source)
  \item \textbf{Timeframe:} 2023-09-01T00:00:00Z (inclusive) to 2023-09-02T00:00:00Z (exclusive)
  \item \textbf{Configuration:} \texttt{source\_type="CSV"}, \texttt{record\_format="CSV"} (overrides default JSON), 24-hour ingestion window, end-exclusive.
  \item \textbf{Context:} First step of an 8-tool compliance pipeline; no prior environment state.
\end{itemize}
\vspace{2mm}

\textbf{Step 2: Agent Mini-task Execution.}
\begin{itemize}[leftmargin=1.5em, itemsep=3pt]
  \item \textbf{Agent:} \texttt{TradeRecordIngestor(source\_type="CSV", source\_identifier="file\_2023\_09.csv", start\_time="2023-09-01T00:00:00Z", end\_time="2023-09-02T00:00:00Z", record\_format="CSV")}
  \item \textbf{Tool:} \texttt{\{"status\_code": 200, "response": \{"raw\_records": [\{"timestamp": "2023-09-01T01:15:00Z", "side": "BUY", "qty": 100, "symbol": "EURUSD"\}, \{"timestamp": "2023-09-01T12:30:00Z", "side": "SELL", "qty": 50, "symbol": "GBPUSD"\}]\}\}}
\end{itemize}
\vspace{2mm}

\textbf{Step 3: Ground-Truth Validation.}
\begin{itemize}[leftmargin=1.5em, itemsep=2pt]
  \item \textbf{Tool Call Equality:} True
  \item \textbf{Task Solved:} True (status 200, all required parameters present, types match schema)
  \item \textbf{Explanation:} The agent's emitted call matches the expected ground-truth call parameter-for-parameter.
\end{itemize}
\vspace{2mm}

\textbf{Step 4: Environment Update.}
\begin{itemize}[leftmargin=1.5em, itemsep=2pt]
  \item \textbf{Source:} \texttt{file\_2023\_09.csv} (CSV, 2023-09-01 window)
  \item \textbf{Ingested Records:} 2023-09-01T01:15:00Z --- BUY 100 EURUSD; 2023-09-01T12:30:00Z --- SELL 50 GBPUSD
  \item \textbf{Downstream:} Output passed to next tool \texttt{RecordValidator} for schema check against \texttt{schema\_version="v1.0"}.
\end{itemize}
}
\end{tcolorbox}

\begin{tcolorbox}[colback=orangePPT!5!white,
                  colframe=orangePPT,
                  title= Task Trajectory,
                  fonttitle=\bfseries,
                  enhanced,
                  breakable,
                  segmentation style={draw=none}]
{\small \raggedright
\textbf{Initial Task.} Aggregate, archive, and maintain trade records to satisfy statutory record-keeping and retention requirements. Ingest CSV trade records from \texttt{file\_2023\_09.csv} for the window 2023-09-01T00:00:00Z to 2023-09-02T00:00:00Z, validate them against schema \texttt{v1.0}, normalize to the \texttt{INTERNAL\_V1} schema, and tag each with metadata (\texttt{source\_system}, \texttt{ingestion\_timestamp}, \texttt{jurisdiction=US}, \texttt{record\_type=execution}). Resolve the statutory retention expiry for US FX execution records under regulator CFTC. Encrypt the tagged records with the Base64-encoded 256-bit key \texttt{MDEyMzQ1Njc4OWFiY2RlZjAxMjM0NTY3ODlhYmNkZWY=} and archive them to \texttt{s3://secure-archive/trades/2023-09/} in zip format with access granted to \texttt{alice} and \texttt{bob}. Finally, verify retention compliance against CFTC requirements.
\vspace{1mm}
\hrule
\vspace{2mm}
\begin{itemize}[leftmargin=1.5em, itemsep=3pt]
\item \textbf{Agent:} \texttt{TradeRecordIngestor(source\_type="CSV", source\_identifier="file\_2023\_09.csv", start\_time="2023-09-01T00:00:00Z", end\_time="2023-09-02T00:00:00Z", record\_format="CSV")}
\item \textbf{Tool:} \texttt{\{"status\_code": 200, "response": \{"raw\_records": ["2023-09-01T01:15:00Z,BUY,100,EURUSD", "2023-09-01T12:30:00Z,SELL,50,GBPUSD"]\}\}}
\item \textbf{Agent:} \texttt{RecordValidator(raw\_records=[\dots], schema\_version="v1.0")}
\item \textbf{Tool:} \texttt{\{"status\_code": 200, "response": \{"valid\_records": [\dots 2 records\dots], "invalid\_records": []\}\}}
\item \textbf{Agent:} \texttt{SchemaNormalizer(valid\_records=[\dots], target\_schema="INTERNAL\_V1")}
\item \textbf{Tool:} \texttt{\{"status\_code": 200, "response": \{"normalized\_records": ["2023-09-01T01:15:00Z|BUY|100|EURUSD", "2023-09-01T12:30:00Z|SELL|50|GBPUSD"]\}\}}
\item[] \hspace{2em}$$\vdots$$
\item \textbf{Agent:} \texttt{RecordEncryptor(tagged\_records=[\dots], encryption\_key="MDEyMzQ1\dots", algorithm="AES-256")}
\item \textbf{Tool:} \texttt{\{"status\_code": 400, "response": \{"error": "IV required for AES-256-CBC mode but not provided."\}\}}
\item \textbf{Agent:} \texttt{RecordEncryptor(tagged\_records=[\dots], encryption\_key="MDEyMzQ1\dots", algorithm="AES-256-CBC", iv="MDAwMTAyMDMwNDA1MDYwNw==")}
\item \textbf{Tool:} \texttt{\{"status\_code": 200, "response": \{"encrypted\_records": ["c2FtcGxlY3lwaGVydGV4dDE=", "c2FtcGxlY3lwaGVydGV4dDI="]\}\}}
\item \textbf{Agent:} \texttt{SecureArchiveWriter(encrypted\_records=[\dots], storage\_location="s3://secure-archive/trades/2023-09/", retention\_expiry\_date="2028-09-01T00:00:00Z", access\_control\_list=["alice", "bob"], archive\_format="zip")}
\item \textbf{Tool:} \texttt{\{"status\_code": 201, "response": \{"archive\_id": "arch-9f7b3c2d-4e1a-4d9f-8a6b-2c5e7f1a9b0c"\}\}}
\item \textbf{Agent:} \texttt{RetentionComplianceChecker(archive\_id="arch-9f7b3c2d-4e1a-4d9f-8a6b-2c5e7f1a9b0c", regulator="CFTC", jurisdiction="US")}
\item \textbf{Tool:} \texttt{\{"status\_code": 200, "response": \{"compliant": true, "retention\_expiry": "2028-09-01T00:00:00Z", "checks\_passed": ["retention\_window", "encryption\_at\_rest", "access\_control"]\}\}}
\end{itemize}
}
\end{tcolorbox}

\section{Tool Distribution}
\label{sec:dataset_details}
\begin{figure}[H]
    \centering
    \includegraphics[width=\linewidth]{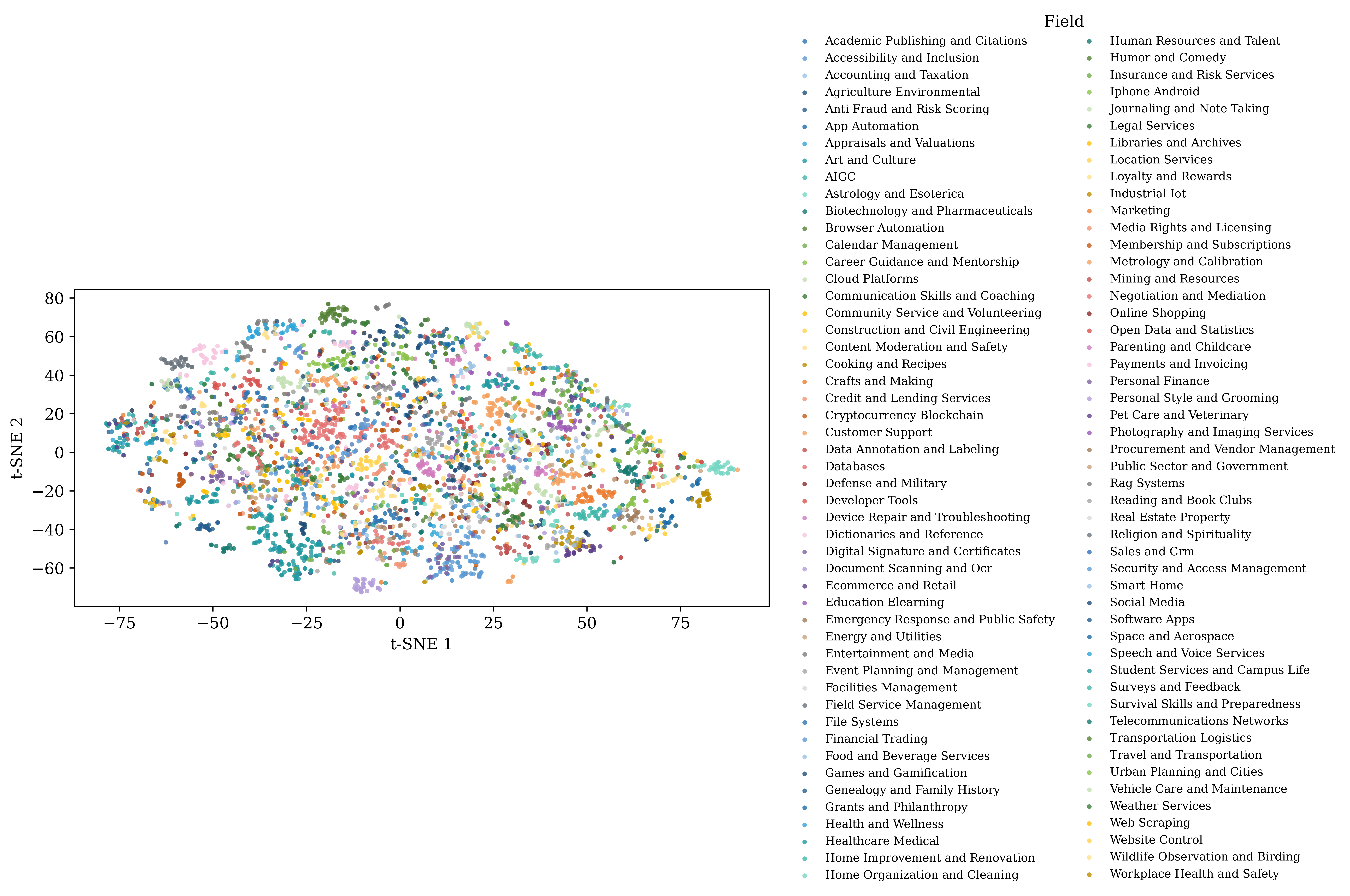}   
    \caption{ Distribution of tool embeddings across 100 fields } 
    \label{fig:full_fields}
\end{figure}

Here, Figure~\ref{fig:full_fields} shows that even at 100 fields, the tools remain diverse.  

\newpage

\end{document}